\documentclass[a4paper]{article}
\usepackage{spconf,amsmath,graphicx}
\usepackage{multirow}
\usepackage{booktabs}
\usepackage{url}
\usepackage{color}

\title{A CROSS-CORPUS STUDY ON SPEECH EMOTION RECOGNITION}
\name{Rosanna Milner$^1$, Md Asif Jalal$^1$, Raymond W. M. Ng$^2$, Thomas Hain$^1$}
\address{
  $^1$University of Sheffield, UK and $^2$Emotech Labs, UK}

\begin{document}
\maketitle

\begin{abstract}

For speech emotion datasets, it has been difficult to acquire large quantities of reliable data and acted emotions may be over the top compared to less expressive emotions displayed in everyday life.
Lately, larger datasets with natural emotions have been created. Instead of ignoring smaller, acted datasets, this study investigates whether information learnt from acted emotions is useful for detecting natural emotions.
Cross-corpus research has mostly considered cross-lingual and even cross-age datasets, and difficulties arise from different methods of annotating emotions causing a drop in performance.
To be consistent, four adult English datasets covering acted, elicited and natural emotions are considered.
A state-of-the-art model is proposed to accurately investigate the degradation of performance. The system involves a bi-directional LSTM with an attention mechanism to classify emotions across datasets. 
Experiments study the effects of training models in a cross-corpus and multi-domain fashion and results show the transfer of information is not successful. 
Out-of-domain models, followed by adapting to the missing dataset, and domain adversarial training (DAT) are shown to be more suitable to generalising to emotions across datasets.
This shows positive information transfer from acted datasets to those with more natural emotions and the benefits from training on different corpora.

\end{abstract}
\noindent\textbf{Index Terms}: speech emotion recognition, cross-corpus, bi-directional LSTM, attention, domain adversarial training

\section{INTRODUCTION}

Speech emotion recognition (SER) aims to automatically detect human emotional states from audio \cite{Dellaert1996,Picard1997} and many reviews into methods, features and datasets exist \cite{Koolagudi2012,Ayadi2011,Anagnostopoulos2015,Ververidis2006}. The problem is complex because emotion cannot be clearly defined, let alone accurately detected. This was initially analysed in the field of psychology where various tools to represent human emotion were studied, such as Plutchik's wheel of emotion \cite{Plutchik1997} and the hourglass of emotions \cite{Cambria2011}. In SER, it is common to adopt classification schema similar to Ekmans's `big-six' emotions \cite{Ekman1992}, which are \textit{happy}, \textit{sad}, \textit{anger}, \textit{surprise}, \textit{disgust} and \textit{fear}.

Many SER studies carry out investigations on specially designed datasets in which human voice is recorded and emotions are annotated. This is challenging as many datasets vary in terms of language, age, labelling scheme (categorical or dimensional), emotions annotated and how the emotion was produced (acted, elicited or simulated, or natural). 
A lot of research within SER has focussed on a single dataset avoiding different annotation problems \cite{Atcheson2018}. Cross-corpus has been investigated in the past with initial work by Schuller et al \cite{Schuller2010} using various normalising schemes across six datasets.
In \cite{Hassan2013}, to relieve the feature distribution mismatch between training and test speech, importance weights are learnt for a support vector machine (SVM) which applies three domain adaption methods for children's speech corpora. Alternatively, \cite{Deng2013,Deng2014,Deng2017} learnt new representations for training and test speech via an autoencorder based domain adaptation framework. The domain-adaptive least squares regression model (DALSR) \cite{Zong2016,Zong2016a} calculates the maximum mean discrepancy (MMD) to balance the feature distribution difference using transfer non-negative matrix factorisation (TNNMF). Kim et al. \cite{KimIS2017} used auxiliary gender and naturalness recognition tasks in a multi-task learning setting for emotion recognition. Most recently, to deal with the unsupervised problem, the domain-adaptive subspace learning (DoSL) approach was proposed where an SVM is trained based on labelled training set speech signals \cite{Liu2018}. 

However, the majority of this cross-corpus work has been cross-lingual and occasionally both cross-lingual and cross-age. In the past this has been arguably necessary given the sparsity of emotional databases. But now with the introduction of larger datasets, such as the MOSEI database which has around 65 hours of natural emotion data \cite{mosei}, more effort can be put on cross-corpus research focussing on a single language and a single age group.  
Additionally, more natural datasets allow research to move away from unrealistic acted emotions. However, given the number of emotional datasets that have already been created, acted or not, it is interesting to investigate whether information can be learnt from these datasets to improve performance on others.

Deep neural networks, such as long short-term memory (LSTM) networks \cite{Hochreiter1997} and attention mechanism \cite{Bahdanau2014}, were employed in various emotion recognition studies \cite{KahouICML2013,Zong2016a,KimIS2017,Beard2018}. 
Recently, in speech recognition, domain adversarial training (DAT) was proposed as a neural network adaptation approach to combat the undesirable variability across different data distributions \cite{GaninJMLR2016}. DAT is applicable to the SER task \cite{Abdel-Wahab2018}, and achieves representation learning even if the training and test domains are unknown. Specifically, a gradient reversal layer is applied in the domain classifier in an attempt to bring the representations closer. Training models using DAT not only saves time, by learning two representations simultaneously, it also aims to improve the representations learnt by using the transfer of information between the two tasks.

This work considers cross-corpus SER within the same language (English) and same age (adult) and investigates what can be learnt between acted and natural datasets, the first study of this kind. A bidirectional-LSTM with an attention mechanism approach is considered for training the cross-corpus models. The information transfer to other datasets is investigated and further experiments looking into domain adversarial training (DAT) and out-of-domain (OOD) models, then adapting these models, are considered.
The approach is described in Section \ref{sec:app}, the datasets studied are detailed in Section \ref{sec:data}, the experimental setup and results are shown in Section \ref{sec:exp}, Section \ref{sec:dis} contains the discussion and lastly, the conclusions are stated in Section \ref{sec:conc}.

\section{APPROACH}
\label{sec:app}
The proposed system is derived from the triple attention network described in \cite{Beard2018} which is based on work by Nam et al \cite{Nam2017}. Both these systems use more than one modality (audio, visual and textual), whereas this work focusses on audio only. The network consists of an encoder which contains a bi-directional LSTM (BLSTM) followed by an attention mechanism and an emotion classifier. For DAT experiments, a domain classifier for corpus ID is included as well.

\subsection{Bi-directional LSTM (BLSTM)}
LSTM networks rely on the temporal order of the sequence, ignoring the future context. BLSTMs \cite{Graves2013} introduce a second layer of hidden connections flowing in the opposite temporal order as a method to exploit the contextual information from the past and the future \cite{Schuster1997}. 
The output vector from the BLSTM for a sequence of length $T$  contains the temporal representation of the sequence from time step $1$ to $T-1$. When considering the final time step, the overall prediction across the sequence is acquired. However, considering all the time steps, a temporal feature distribution over the sequence is obtained which is useful for SER tasks.

\subsection{Attention Mechanism}
Attention mechanism has the flexibility of computing long-term inter-sequence dependencies. In this work, the attention value on the overall temporal feature space from the BLSTM is computed. The attention mechanism focuses the network onto specific parts of itself by computing the global mean which captures global information. The global mean is multiplied over the whole temporal vector to compute the positional dependency of each element with $tanh$ non-linearity. The resulting vector is used to compute the attention weights using $softmax$ scoring.
The soft attention mechanism is adopted for this work and the multiplicative method is applied as in \cite{Beard2018}, similar to \cite{Nam2017}, as the authors found their results similar to the standard concatenated approach \cite{Bahdanau2014}. The main challenge for cross corpus SER is to minimise the long-term spatio-temporal variation between different corpora for a given categorical distribution and one solution is to model global spatio-temporal dependencies between corpora. The proposed framework with attention modelling is an efficient way to approach the problem.

\subsection{Emotion Classifier}
The predictor stage of the network contains a fully connected linear layer which projects the attention output down to the number of emotions present. This is then passed through a $softmax$ layer before computing the loss.

\subsection{Domain Classifier: DAT}

Domain adversarial training (DAT) is a neural network training technique which incorporates domain adaptation into  the process of learning representation \cite{GaninJMLR2016}. Let $i$ denote the index of the training samples, $y$ denotes the label and $d$ denotes the domain, the SER task is formulated as multitask learning with the primary task being emotion classification, with loss function $L^{i}_{y}$, and domain prediction being the auxiliary task, with loss function $L^{i}_{d}$. DAT models are trained in parallel with the aim to minimise the loss of the target label predictor (emotion category) and the loss of the domain label predictor (corpus ID). Following other DAT studies, it is assumed that the source of the data is an important indication of domain and so corpus ID is used as the domain label indicator. The network contains an arbitrary number of shared layers where their parameters contribute to both target and domain prediction. These shared layers provide high-level, regularised representation. Like most settings in multi-task learning, the network then branches off into the target and domain classifiers. The objective of the network is:
%
\begin{equation}\label{eqn:dat}
E(\theta_f,\theta_y,\theta_d) = \dfrac{1}{n} \sum^{n}_{i=1} L^{i}_{y}(\theta_f,\theta_y) - \lambda \Big(\dfrac{1}{n}\sum^{n}_{i=1} L^{i}_{d}(\theta_f,\theta_d)\Big)
\end{equation}
which is minimised w.r.t. $f$ and $y$ but maximised w.r.t. $d$. The parameters of the shared layers are represented by $\theta_f$, and $\theta_y$ and $\theta_d$ represent the parameters of the target and domain classifier layer respectively. The number of training samples is referred to as $n$. To implement Equation \ref{eqn:dat}, the first domain classifier layer is implemented as a gradient reversal layer. The gradients being back-propagated from the domain layer are reversed multiplying by a negative factor. The $\lambda$ controls how optimisation is biased to the target or the domain.

\section{Datasets}
\label{sec:data}

This work only considers English speaking adult datasets across three emotion production types: two acted datasets, eNTERFACE \cite{enterface} and  RAVDESS \cite{ravdess}, one elicited dataset, IEMOCAP \cite{iemocap}, and one natural dataset, MOSEI \cite{mosei}. Table \ref{tab:emotions} gives an overview of the emotion classes covered in each dataset.

\begin{table}[t!]
\centering
\begin{tabular}{lrrrr}
\toprule
Emotion & ENT & RAV & IEM  & MOS \\
\midrule
\textit{happy}	& 212 &	192 &  595 & 12503 \\
\textit{sad}	    & 215 &	192 & 1084 &  7181 \\
\textit{anger}	    & 215 &	192 & 1103 &  7052 \\
\textit{surprise}    & 215 & 192 &  107 &  2592 \\
\textit{disgust}     & 215 & 192 &    2 &  5436 \\
\textit{fear}        & 215 & 192 &   40 &  2128 \\
\textit{neutral}     & -   &  96 & 1708 & (1239)* \\
\textit{frustration} & -   & -   & 1849 &  -     \\
\textit{excitement}  & -   & -   & 1041 &  -     \\
\textit{other}       & -   & -   &    3 &  -     \\
\textit{calm}        & -   & 192 & -    &  -     \\
\midrule
Total	    & 1287 & 1440 & 7532 & 10441* \\
\bottomrule
\end{tabular}
\caption{Number of segments across the emotions in each of the four datasets. *MOSEI contains the emotion \textit{neutral} indirectly, in the form of a reference vector of zeros, and has 10441 multi-labelled segments with 24677 emotion labels.
	}
\label{tab:emotions}
\end{table}

\subsection{eNTERFACE}

For eNTERFACE (ENT), Project 2 from 2005 has the big-six emotion classes and contains about 1 hour of acted utterrances \cite{enterface}. There are 5 recordings of each emotion for every speaker. There are 44 speakers (8 female), however due to further inspection of the data this was reduced to 43 (Spkr6's recordings have not been segmented) and 1287 segments (Spkr23 is missing three \textit{happy} segments).
The speakers are from 14 nations, not all native speakers, resulting in different English accents.
A training set is created with 38 speakers (Spkr1 to Spkr39 without Spkr6) and the test set contains the last 5 speakers (Spkr40 to Spkr44).

\subsection{RAVDESS}

RAVDESS (RAV) is another acted dataset containing 24 speakers (12 male and 12 female) with 1.5 hours of speech \cite{ravdess}. The dataset contains the big-six emotions and also \textit{calm} and \textit{neutral}. 
It contains North American English accented speech.
The data is split into the first 19 speakers for training and the last 5 for testing. 
The reference labels the emotions as 0 for non-existing emotion, 1 as exists and 2 as strong emotion. All 2 values have been changed to 1 for classification.

\subsection{IEMOCAP}
\label{sec:iemocap}

IEMOCAP (IEM) contains utterances from 10 speakers (5 male and 5 female) over 12 hours \cite{iemocap}. There are five dyadic sessions (between two speakers) which are either scripted or improvised to elicit emotions. As well as the big-six emotions, it also contains other categories: \textit{excitement}, \textit{frustration}, \textit{neutral} and \textit{other}. 
The English spoken has a North American accent. 
The first four sessions, containing 8 speakers, are used as training data and the last session, containing 2 speakers, is used for testing. 
In the literature it is common for IEMOCAP to be evaluated as four classes: \textit{happy}, \textit{sad}, \textit{anger} and \textit{neutral} (where \textit{excitement} is combined with \textit{happy} to give 1636 \textit{happy} segments across all the sessions) \cite{Li2018}. This test set will be referred to as IEM4.

\subsection{MOSEI}

MOSEI (MOS) is currently the largest sentiment and emotion dataset at around 65 hours of data and more than 1000 speakers \cite{mosei}. Utterances have been segmented from YouTube videos and annotated for the big-six emotions using Amazon Mechanical Turk. As data is collected from YouTube and the videos are not specifically designed as an emotion dataset, the emotional speech is seen as natural. 
The videos contain English speech but it is unclear which accents are covered. 
The official training, validation and test splits\footnote{\url{https://github.com/A2Zadeh/CMU-MultimodalSDK/blob/master/mmsdk/mmdatasdk/dataset/standard_datasets/CMU_MOSEI/cmu_mosei_std_folds.py}} for the ACL 2018 conference have been used, where the training and validation sets are combined for training. 
The emotion labels for classification, which are in the range 0 to 3, have been changed to binary values, whether the emotion exists or not. 

\section{Experiments}
\label{sec:exp}

Several different experiments are performed to study the effects of cross-corpus SER. Firstly, a comparison of the efficacy of different features in SER is carried out. Secondly, cross-corpus experiments are performed to see the effectiveness of SER using matched and unmatched data. This is followed by training a model on all the domains for comparison, referred to as multidomain, MD. Next, out-of-domain, OOD, and adaptation is investigated as well as a system with domain adversarial training, DAT.

\subsection{Experimental setup}

\begin{table}[t!]
	\centering
	\begin{tabular}{lcccccc}
		\toprule
		Split & ENT & RAV & IEM & IEM4  & MOS & Total \\
		\midrule
		Train & 1137 & 912& 2345 & -  & 7193   & 11587\\
		\textit{hours} & \textit{0.6} & \textit{0.9}& \textit{3.1} & \textit{-}  & \textit{13.7}   & \textit{18.3}\\
		\midrule
		Test & 150 & 240& 586 & 1241   & 2009   & 4226\\
		\bottomrule
	\end{tabular}
	\caption{For each of the four datasets, the number of training and testing segments are shown, where IEM4 is testing only. 
	}
	\label{tab:splits}

\end{table}

The system is implemented in PyTorch \cite{pytorch}. 
The BLSTM contains two hidden layers of 512 nodes each. The output layer of size 1024 feeds in to the attention mechanism computing a context vector of size 128 which is projected to 1024 nodes. This is then passed to the predictor stage which linearly projects to the number of classes. The cross entropy loss function is applied which is preceded by a $softmax$ layer in the PyTorch implementation. 
The Adam optimiser \cite{adamoptim} is used  with the initial learning rate of 0.0001. As Adam adaptively optimises the learning rate but does not change it, the PyTorch approach of ReduceLROnPlateau was investigated. The optimum patience setting was found to be 4 epochs with a multiplicative factor of 0.8. The models were trained to 200 epochs and the best model chosen by averaging the results across the datasets as adding a stopping criterion could influence the effect on the different datasets. Further work is necessary to find the optimum stopping criterion which does not give bias or priority towards one particular dataset. The training batch size was set to 1 as various batch sizes were tried fixing the segment lengths but no significant difference of performance was found. 
For the DAT experiment, the optimum negative variable $\lambda$ was found to be 0.007.

Table \ref{tab:splits} shows the training and testing data splits for each dataset where only the big-six emotions are considered. For eNTERFACE and RAVDESS, the SHoUT speech activity detector \cite{shout} has been applied to remove the start and end silences around the speech as exact timing was not provided.

\subsection{Evaluation} 

The metrics used to evaluate the approach are the unweighted accuracy (UA) and the weighted accuracy (WA). The UA calculates accuracy in terms of the total correct predictions divided by total samples, which gives the same weight to each class:
\begin{equation}
UA=\dfrac{TP+TN}{P+N}
\end{equation}
where $P$ is the number of correct positive instances (equivalent to $TP+FN$) and $N$ is the number of correct negative instances (equivalent to $TN+FP$).
As some of the datasets are imbalanced across the emotion classes, see Table \ref{tab:emotions}, the WA is calculated which weighs each class according to the number of samples in that class:
\begin{equation}
WA=\dfrac{1}{2}(\dfrac{TP}{P}+\dfrac{TN}{N})
\end{equation}

\subsection{Baseline systems}
\label{sec:baseline}
Baseline system results from the literature are difficult to find due to the fact that this work requires the big-six emotions and many datasets contain more, or less, and therefore evaluation results are not equivalent. Additionally, not all research states both unweighted accuracy and weighted accuracy.
For ENT, a system which applies a convolutional recurrent neural network (CRNN) with an attention mechanism \cite{Huang2017} achieves a UA of 91.7\%.
For 6-class classification, RAV and IEM are not typically evaluated this way, IEM results focus on the four classes (IEM4) mentioned in Section \ref{sec:iemocap}.
The best UA found for the IEM4 test set is 71.8\% \cite{Li2018} presenting an attention pooling based representation learning method.
A cross-corpus UA of 64.0\%  is shown in \cite{Luo2018} where a two-channel system approach is adopted combining high-level statistic features with a CRNN. For WA, 56.1\% is shown in \cite{Desplanques2018} which also presents a cross-corpus WA of 48.4\% applying factor analysis to find emotion factors to classify.
For MOS, a WA of 52.5\% is presented in \cite{Beard2018} which this work is based on.

\begin{table}[t!]
	\centering
	\begin{tabular}{lccc}
		\toprule
		Features	& \#dimensions & UA\%  &  WA\% \\
		\midrule
		LMFB	 & 20 & 94.4 & 90.0 \\
		LMFB     & 23 & \textbf{95.6} & \textbf{92.0} \\
		LMFB     & 40 & 92.7 & 86.8 \\
		LMFB     & 60 & 94.4 & 90.0 \\
		LMFB     & 80 & 94.9 & 90.8 \\
		PLP      & 14 & 90.7 & 83.2 \\
		MFCC     & 13 & 87.8 & 78.0 \\
		COVAREP  & 74 & 84.4 & 72.0 \\
		\bottomrule
	\end{tabular}
	\caption{Performance for ENT data where models have been trained using different acoustic features where bold shows the best performance. }
	\label{tab:features}
\end{table}

\subsection{Results: Acoustic features}

For deep learning, it has been shown that log-Mel filterbanks (LMFB) yield better performance over Mel frequency cepstral coefficients (MFCCs) \cite{Hermansky1998}.
However, SER has seen better improvements when considering other acoustic features so different types are investigated, such as MFCC, PLP (perceptual linear prediction), and COVAREP \cite{Degottex2014} features. COVAREP features, which include pitch, were extracted applying the \textit{COVAREP\_feature\_extraction.m} script and the rest were extracted using the HTK toolkit \cite{htkbook}.
The eNTERFACE dataset is used for investigating the training setup as it is the smallest dataset in terms of time and has a high number of speakers with almost completely balanced emotions. 
Table \ref{tab:features} displays the UA and WA results across the different features extracted. 
LMFBs with 23 dimensions outperform the rest in terms of UA and WA, therefore the remaining experiments are performed with 23 dimensional LMFBs.

\begin{table*}[t!]
	\centering
	\begin{tabular}{llcccccccccccc}
		\toprule
		\multirow{2}{*}{Experiment} & Training & & \multicolumn{5}{c}{Unweighted Accuracy, UA\%} & & \multicolumn{5}{c}{Weighted Accuracy, WA\%} \\
		& Data & & ENT & RAV & IEM & IEM4  & MOS  & & ENT & RAV & IEM & IEM4  & MOS  \\
		\midrule
		\multirow{4}{*}{Cross-corpus (CC)} & ENT  	&	& \textbf{95.6*} & 74.7 & 77.9 & 66.2  & 54.8 & & \textbf{92.0} & 54.5 & 56.2 & 58.1 	& 48.5\\
		 & RAV  	&	& 74.7 & \textbf{86.1} & 82.2 & 71.8*  & 56.6 & & 54.4 & \textbf{75.0} & 56.3 & 60.1*  	& 49.4 \\
		 & IEM  	&	& 72.9 & 75.6 & \textbf{90.4} & \textbf{72.0*}  & 66.3 & & 51.2 & 56.2 & \textbf{64.1} & \textbf{67.7*}  & 49.4 \\
		 & MOS  	&	& 78.4 & 73.3 & 82.0 & 70.2 & \textbf{74.5}  & & 61.2 & 52.0 & 54.4 & 58.5 & \textbf{52.8*} \\
		\midrule
		Multi-domain (MD) & all 	&	& 89.8 & 86.1 & 88.8 & 72.0 & 73.3  &  & 81.6 & 75.0 & 64.2 & 66.6  & 55.0  \\
		\midrule
		DAT  & all	&	& 88.7 & 87.1 & 89.3 & 72.1 & 74.1 & & 79.6 & 76.8 & 64.9 & 67.0 & 55.6 \\
		\bottomrule
	\end{tabular}
	\caption{Cross-corpus (CC), multi-domain (MD) and DAT results across the five testsets, where each row refers to a single model. The asterisks refer to the results which can be compared to baseline systems and bold refers to the matched condition.}
	\label{tab:crosscorpus}
\end{table*}

\begin{table*}[t!]
	\centering
	\begin{tabular}{llcccccccccccc}
		\toprule
		\multirow{2}{*}{Experiment} & \multirow{2}{*}{Training Data} & & \multicolumn{5}{c}{Unweighted Accuracy, UA\%} & & \multicolumn{5}{c}{Weighted Accuracy, WA\%} \\
&	 & & ENT & RAV & IEM & IEM4  & MOS  & & ENT & RAV & IEM & IEM4  & MOS  \\
		\midrule
		Out-of-domain 	& three, mismatched & & 75.6 & 73.3 & 83.7 & 69.7 & 61.4 & & 56.0 & 52.0 & 58.0 & 59.4 & 48.1 \\
		+adaptation	& matched  & & 91.6 & 85.1 & 90.0 & 72.0 & 69.3 & & 84.8 & 73.2 & 67.1 & 66.9 & 55.5 \\
		\bottomrule
	\end{tabular}
	\caption{Out-of-domain (OOD) and adaptation (+adaptation) results across the five testsets. There are two models for each dataset, OOD and  OOD with adaptation.
		}
	\label{tab:ood}
\end{table*}

\subsection{Results: Cross-corpus}

The cross-corpus (CC) results are seen in Table \ref{tab:crosscorpus} where the performances in the bold diagonal refer to the matched condition (trained and tested on the same dataset) and the non-bold performances are the mismatched condition (trained and tested on different datasets). Each row in the table refers to a single model.
In terms of the baseline systems described in Section \ref{sec:baseline}, asterisks are displayed next to comparable results. The cross-corpus and matched results achieve better results than the baselines. This shows the proposed system is set up well for recognising emotions from speech and acceptable to use for this study.
The best performance is found in the matched condition as would be expected. 
The model trained on IEM achieves the best mismatched results for RAV and MOS testsets but not for ENT. As an elicited dataset, it can be argued to be close to both the acted and more natural scenarios. 
Comparing the models trained on the acted datasets, it can be seen that the RAV model performs better for IEM, IEM4 and MOS than the ENT model. This shows that despite both being similar acted datasets, there seems to be more useful information to learn from a model trained on RAV than on ENT. This reinforces the challenges from datasets with different annotations \cite{Atcheson2018}.
Performance reduces when moving to the natural dataset MOS, showing the difficulty of classifying more naturally produced emotions.

\subsection{Results: Multi-domain}

The multi-domain (MD) model is trained on the four datasets and results are shown in Table \ref{tab:crosscorpus}. 
The model does not outperform the cross-corpus matched results, except in one case. 
The 2.2\% improvement in the WA performance for MOS begins to show how training on acted datasets could be beneficial for some natural datasets.
However, the multi-domain model does consistently outperform the best cross-corpus mismatched results. For UA, the average improvement between the best mismatched model and the multi-domain model is 7.1\%, whereas for WA this is larger at 11.8\%. The gains are larger for ENT and RAV, the smaller acted datasets, than the IEM, IEM4 and MOS testsets. It is known that neural networks can improve performance with more data, so this shows the benefit of including different corpora, particularly for smaller datasets, despite the annotation challenges. Information about the emotions in different settings is shown to be learnt and beneficial across datasets.

\begin{table*}[th!]
	\centering
	\begin{tabular}{llccccccccccccccc}
		\toprule
		\multirow{2}{*}{Exp.} & Training & \multicolumn{7}{c}{Unweighted Accuracy, UA\%} & & \multicolumn{7}{c}{Weighted Accuracy, WA\%} \\
		& Data &  hap. & sad & ang. & sur.  & dis. & fear & neu.  & & hap. & sad & ang. & sur.  & dis. & fear & neu.  \\
		\midrule
		\multirow{4}{*}{CC} & ENT 	& \textbf{74.1} & 67.4 & 70.8 & 74.9 & 78.3 & 86.9 & 61.3 & & 58.4 & 65.8 & \textbf{69.0} & \textbf{61.4} & \textbf{59.3} & \textbf{57.9} & \textbf{59.1} \\
		& RAV 	& 61.5 & 77.3 & 71.3 & 87.2 & 71.1 & 83.0 & \textbf{69.1} & & 58.7 & 65.8 & 61.3 & 57.4 & 55.2 & 55.4 & 50.0 \\
		& IEM 	& 66.4 & \textbf{77.4} & 56.8 & 87.8 & \textbf{85.6} & \textbf{88.4} & \textbf{69.1} & & 53.3 & \textbf{68.1} & 67.8 & 50.0 & 50.0 & 50.0 & 50.0 \\
		& MOS  & 60.7 & 67.3 & \textbf{75.8} & \textbf{88.1} & 84.7 & \textbf{88.4} & 66.3 & & \textbf{58.9} & 66.6 & 53.6 & 49.9 & 51.6 & 50.0 & 55.4 \\
		\midrule
		DAT 	& all & 75.0 & 81.5 & 81.4 & 89.2 & 86.5 & 90.4 & 69.0 & & 67.4 & 76.0 & 78.0 & 64.5 & 70.3 & 57.5 & 53.1 \\
		\bottomrule
	\end{tabular}
	\caption{Performances across the emotions for the cross-corpus (CC) and DAT models, where each row refers to a single model. Results in bold refer to the best performances for each emotion across the CC models.}
	\label{tab:7classes}
\end{table*}

\subsection{Results: DAT}

The DAT model is trained on the four datasets and performance is displayed in Table \ref{tab:crosscorpus}.
Comparing to the multi-domain model, the DAT model performs better in all testsets except for ENT. This shows, in most cases, how the domain information can help models to generalise for recognising emotions across datasets.  
In the ENT dataset, each emotion has five defined sentences which are spoken by every speaker. This high degree of matching between the training and test sets contributes to the good performance when training and testing on ENT. However, this matching property cannot be picked up effectively by the domain classifier.
There are some small gains in comparison to the cross-corpus matched condition results but it is not consistent across the datasets or the two accuracies. Nevertheless, the DAT method outperforms the best mismatched results, as the multi-domain model does, reinforcing that data from different datasets can be useful for SER.

\subsection{Results: Out-of-domain and adaptation}

The OOD results with the corresponding adaptation performance are displayed in Table \ref{tab:ood}. There are four OOD models, each trained on three datasets. The performance for each testset shown comes from the mismatched trained model. Results for IEM and IEM4 are from the same model. 
The OOD models are then adapted to the dataset previously not included and this matched result is shown.
The OOD results do not reach the accuracies of the multi-domain model, but they do outperform the worst cross-corpus mismatched result. This shows that learning on additional data is useful even if it's from different, mismatched datasets.
When adapting the OOD models, large gains are seen reaching similar results as the multi-domain and DAT models. The average UA improvement is 8.9\% and 14.8\% for WA. This shows how important it is to train on matched data.

\subsection{Results: Emotion classes}
\label{sec:dis}

The experiments have shown how acted datasets can benefit more natural datasets with the multi-domain, DAT and OOD with adaptation models all achieving WA results outperforming the cross-corpus matched model for MOS.
Complimentary to looking at the performances across datasets, comparing the performances between the emotion classes can give insight into how the models are behaving. The cross-corpus and DAT models are considered, and the results are displayed in Table \ref{tab:7classes} where the \textit{neutral} performance comes from only one testset, IEM4.
For the cross-corpus results, the bold numbers refer to the best performance for that emotion. For the UA results, it is fairly inconsistent which model is better, however the models trained on IEM and MOS give the better performances for all the emotions except for \textit{happy}. 
%
For the WA results, it shows the opposite, that training on ENT is better across five of the seven emotions. This seems to suggest that acted datasets are good at distinguishing emotions, even across datasets with different emotion production types.
The DAT model gives better performance across all emotions in terms of UA and WA, except for the weighted accuracy of \textit{fear} and \textit{neutral} emotions.

\section{Conclusions}
\label{sec:conc}


This study has carried out an investigation into cross-corpus speech emotion recognition. It is the first study of its kind investigating the degradation of performance when moving from acted to more natural datasets.
Initial experiments into the optimum acoustic features for this set-up showed that LMFB features outperform MFCCs, PLPs and COVAREP features.
Cross-corpus experiments show the matched results outperform mismatched and that the model trained on the elicited dataset achieves best mismatched performance in most cases.
In the multi-domain setting, the model outperforms the best mismatched results showing more data is beneficial, even if the emotions are in a different setting or production type.
The DAT experiment shows how including the domain information does help the model to generalise and outperform the multi-domain model for all but one dataset.
The OOD models show that learning from multiple mismatched datasets is more useful than training on a single mismatched dataset, and adapting gives large improvements showing the importance of training on matched data. 
When looking across the emotion class performance, ENT data shows to be best in terms of WA for most of the emotions showing acted datasets may have their uses for the more natural datasets.

\newpage
\clearpage

\bibliographystyle{IEEEbib}

\bibliography{biblio}

\begin{thebibliography}{10}

\bibitem{Dellaert1996}
Frank Dellaert, Thomas Polzin, and Alex Waibel,
\newblock ``Recognizing emotion in speech,''
\newblock in {\em Spoken Language Processing, Philadelphia, PA, USA, Oct 3-6},
  1996.

\bibitem{Picard1997}
Rosalind~W. Picard,
\newblock {\em Affective Computing},
\newblock MIT Press, Cambridge, MA, USA, 1997.

\bibitem{Koolagudi2012}
Shashidhar~G. Koolagudi and K.~Sreenivasa Rao,
\newblock ``Emotion recognition from speech: a review,''
\newblock {\em International Journal of Speech Technology}, vol. 15, no. 2, pp.
  99--117, 2012.

\bibitem{Ayadi2011}
Moataz M. H.~El Ayadi, Mohamed~S. Kamel, and Fakhri Karray,
\newblock ``Survey on speech emotion recognition: Features, classification
  schemes, and databases,''
\newblock {\em Pattern Recognition}, vol. 44, no. 3, pp. 572--587, 2011.

\bibitem{Anagnostopoulos2015}
Christos{-}Nikolaos Anagnostopoulos, Theodoros Iliou, and Ioannis Giannoukos,
\newblock ``Features and classifiers for emotion recognition from speech: a
  survey from 2000 to 2011,''
\newblock {\em Artificial Intelligence Review}, vol. 43, no. 2, pp. 155--177,
  2015.

\bibitem{Ververidis2006}
Dimitrios Ververidis and Constantine Kotropoulos,
\newblock ``Emotional speech recognition: Resources, features, and methods,''
\newblock {\em Speech Communication}, vol. 48, no. 9, pp. 1162--1181, 2006.

\bibitem{Plutchik1997}
Robert Plutchik,
\newblock ``Emotion: Theory, research, and experience: Vol. 1. theories of
  emotion,''
\newblock {\em New York: Academic}, 1997.

\bibitem{Cambria2011}
Erik Cambria, Andrew Livingstone, and Amir Hussain,
\newblock ``The hourglass of emotions,''
\newblock in {\em {COST} 2102 Training School}. 2011, Lecture Notes in Computer
  Science, Springer.

\bibitem{Ekman1992}
Paul Ekman,
\newblock ``An argument for basic emotions,''
\newblock {\em Cognition and Emotion}, pp. 169--200, 1992.

\bibitem{Atcheson2018}
Mia Atcheson, Vidhyasaharan Sethu, and Julien Epps,
\newblock ``Demonstrating and modelling systematic time-varying annotator
  disagreement in continuous emotion annotation,''
\newblock in {\em Interspeech, Hyderabad, India, 2-6 Sep}, 2018, pp.
  3668--3672.

\bibitem{Schuller2010}
Bj{\"{o}}rn~W. Schuller, Bogdan Vlasenko, Florian Eyben, Martin W{\"{o}}llmer,
  Andr{\'{e}} Stuhlsatz, Andreas Wendemuth, and Gerhard Rigoll,
\newblock ``Cross-corpus acoustic emotion recognition: Variances and
  strategies,''
\newblock {\em {IEEE} Trans. Affective Computing}, vol. 1, no. 2, pp. 119--131,
  2010.

\bibitem{Hassan2013}
Ali Hassan, Robert~I. Damper, and Mahesan Niranjan,
\newblock ``On acoustic emotion recognition: Compensating for covariate
  shift,''
\newblock {\em {IEEE} Trans. Audio, Speech {\&} Language Processing}, vol. 21,
  no. 7, pp. 1458--1468, 2013.

\bibitem{Deng2013}
Jun Deng, Zixing Zhang, Erik Marchi, and Bj{\"{o}}rn~W. Schuller,
\newblock ``Sparse autoencoder-based feature transfer learning for speech
  emotion recognition,''
\newblock in {\em Humaine Association Conference on Affective Computing and
  Intelligent Interaction, {ACII}, Geneva, Switzerland, Sep 2-5}, 2013, pp.
  511--516.

\bibitem{Deng2014}
Jun Deng, Xinzhou Xu, Zixing Zhang, Sascha Fr{\"{u}}hholz, and Bj{\"{o}}rn~W.
  Schuller,
\newblock ``Universum autoencoder-based domain adaptation for speech emotion
  recognition,''
\newblock {\em {IEEE} Signal Processing Letters}, vol. 24, no. 4, pp. 500--504,
  2017.

\bibitem{Deng2017}
Jun Deng, Xinzhou Xu, Zixing Zhang, Sascha Fr{\"{u}}hholz, and Bj{\"{o}}rn~W.
  Schuller,
\newblock ``Universum autoencoder-based domain adaptation for speech emotion
  recognition,''
\newblock {\em {IEEE} Signal Processing Letters}, 2017.

\bibitem{Zong2016}
Yuan Zong, Wenming Zheng, Tong Zhang, and Xiaohua Huang,
\newblock ``Cross-corpus speech emotion recognition based on domain-adaptive
  least-squares regression,''
\newblock {\em {IEEE} Signal Processing Letters}, vol. 23, no. 5, pp. 585--589,
  2016.

\bibitem{Zong2016a}
Yuan Zong, Wenming Zheng, Xiaohua Huang, Keyu Yan, Jingwei Yan, and Tong Zhang,
\newblock ``Emotion recognition in the wild via sparse transductive transfer
  linear discriminant analysis,''
\newblock {\em Journal of Multimodal User Interfaces}, vol. 10, no. 2, pp.
  163--172, 2016.

\bibitem{KimIS2017}
Jaebok Kim, Gwenn Englebienne, Khiet~P. Truong, and Vanessa Evers,
\newblock ``Towards speech emotion recognition "in the wild" using aggregated
  corpora and deep multi-task learning,''
\newblock in {\em Interspeech, Stockholm, Sweden, Aug 20-24}, 2017, pp.
  1113--1117.

\bibitem{Liu2018}
Na~Liu, Yuan Zong, Baofeng Zhang, Li~Liu, Jie Chen, Guoying Zhao, and Junchao
  Zhu,
\newblock ``Unsupervised cross-corpus speech emotion recognition using
  domain-adaptive subspace learning,''
\newblock in {\em {ICASSP}, Calgary, AB, Canada, Apr 15-20}, 2018, pp.
  5144--5148.

\bibitem{mosei}
Amir Zadeh, Paul~Pu Liang, Soujanya Poria, Erik Cambria, and Louis{-}Philippe
  Morency,
\newblock ``Multimodal language analysis in the wild: {CMU-MOSEI} dataset and
  interpretable dynamic fusion graph,''
\newblock in {\em Annual Meeting of the Association for Computational
  Linguistics, {ACL}, Melbourne, Australia, Jul 15-20}, 2018, pp. 2236--2246.

\bibitem{Hochreiter1997}
Sepp Hochreiter and J{\"{u}}rgen Schmidhuber,
\newblock ``Long short-term memory,''
\newblock {\em Neural Computation}, vol. 9, no. 8, pp. 1735--1780, 1997.

\bibitem{Bahdanau2014}
Dzmitry Bahdanau, Kyunghyun Cho, and Yoshua Bengio,
\newblock ``Neural machine translation by jointly learning to align and
  translate,''
\newblock in {\em International Conference on Learning Representations, {ICLR},
  San Diego, CA, USA, May 7-9,}, 2015.

\bibitem{KahouICML2013}
Samira~Ebrahimi Kahou, Christopher~J. Pal, Xavier Bouthillier, Pierre
  Froumenty, {\c{C}}aglar G{\"{u}}l{\c{c}}ehre, and Roland~Memisevic et~al,
\newblock ``Combining modality specific deep neural networks for emotion
  recognition in video,''
\newblock in {\em International Conference on Multimodal Interaction, {ICMI},
  Sydney, NSW, Australia, Dec 9-13}, 2013, pp. 543--550.

\bibitem{Beard2018}
Rory Beard, Ritwik Das, Raymond W.~M. Ng, P.~G.~Keerthana Gopalakrishnan, Luka
  Eerens, Pawel Swietojanski, and Ondrej Miksik,
\newblock ``Multi-modal sequence fusion via recursive attention for emotion
  recognition,''
\newblock in {\em Computational Natural Language Learning, CoNLL, Brussels,
  Belgium, Oct 31 - Nov 1}, 2018, pp. 251--259.

\bibitem{GaninJMLR2016}
Yaroslav Ganin, Evgeniya Ustinova, Hana Ajakan, Pascal Germain, Hugo
  Larochelle, Fran{\c{c}}ois Laviolette, Mario Marchand, and Victor~S.
  Lempitsky,
\newblock ``Domain-adversarial training of neural networks,''
\newblock {\em Journal of Machine Learning Research}, vol. 17, pp. 59:1--59:35,
  2016.

\bibitem{Abdel-Wahab2018}
Mohammed Abdel{-}Wahab and Carlos Busso,
\newblock ``Domain adversarial for acoustic emotion recognition,''
\newblock {\em {IEEE/ACM} Trans. Audio, Speech {\&} Language Processing}, vol.
  26, no. 12, pp. 2423--2435, 2018.

\bibitem{Nam2017}
Hyeonseob Nam, Jung{-}Woo Ha, and Jeonghee Kim,
\newblock ``Dual attention networks for multimodal reasoning and matching,''
\newblock in {\em {IEEE} Conference on Computer Vision and Pattern Recognition,
  {CVPR}, Honolulu, HI, USA, Jul 21-26}, 2017, pp. 2156--2164.

\bibitem{Graves2013}
Alex Graves, Navdeep Jaitly, and Abdel{-}rahman Mohamed,
\newblock ``Hybrid speech recognition with deep bidirectional {LSTM},''
\newblock in {\em {IEEE} Workshop on Automatic Speech Recognition and
  Understanding, Olomouc, Czech Republic, Dec 8-12,}, 2013, pp. 273--278.

\bibitem{Schuster1997}
Mike Schuster and Kuldip~K. Paliwal,
\newblock ``Bidirectional recurrent neural networks,''
\newblock {\em {IEEE} Trans. Signal Processing}, vol. 45, no. 11, pp.
  2673--2681, 1997.

\bibitem{enterface}
Olivier Martin, Irene Kotsia, Beno{\^{\i}}t Macq, and Ioannis Pitas,
\newblock ``The enterface'05 audio-visual emotion database,''
\newblock in {\em International Conference on Data Engineering Workshops,
  {ICDE}, Atlanta, GA, {USA}, Apr 3-7}, 2006, p.~8.

\bibitem{ravdess}
Steven~R. Livingstone and Frank~A. Russo,
\newblock ``{The Ryerson Audio-Visual Database of Emotional Speech and Song
  (RAVDESS)},''
\newblock {\em {PLoS ONE}}, 2018.

\bibitem{iemocap}
Carlos Busso, Murtaza Bulut, Chi{-}Chun Lee, Abe Kazemzadeh, Emily Mower,
  Samuel Kim, Jeannette~N. Chang, Sungbok Lee, and Shrikanth Narayanan,
\newblock ``{IEMOCAP:} interactive emotional dyadic motion capture database,''
\newblock {\em Language Resources and Evaluation}, vol. 42, no. 4, pp.
  335--359, 2008.

\bibitem{Li2018}
Pengcheng Li, Yan Song, Ian~Vince McLoughlin, Wu~Guo, and Lirong Dai,
\newblock ``An attention pooling based representation learning method for
  speech emotion recognition,''
\newblock in {\em Interspeech, Hyderabad, India, 2-6 Sep,}, 2018, pp.
  3087--3091.

\bibitem{pytorch}
Adam~Paszke et~al.,
\newblock ``Automatic differentiation in pytorch,''
\newblock in {\em NIPS-W}, 2017.

\bibitem{adamoptim}
Diederik~P. Kingma and Jimmy Ba,
\newblock ``Adam: {A} method for stochastic optimization,''
\newblock in {\em International Conference on Learning Representations, {ICLR},
  San Diego, CA, USA, May 7-9,}, 2015.

\bibitem{shout}
Marijn Huijbregts,
\newblock {\em Segmentation, diarization and speech transcription : surprise
  data unraveled},
\newblock Ph.D. thesis, University of Twente, Enschede, Netherlands, 2008.

\bibitem{Huang2017}
Che{-}Wei Huang and Shrikanth~S. Narayanan,
\newblock ``Deep convolutional recurrent neural network with attention
  mechanism for robust speech emotion recognition,''
\newblock in {\em {IEEE} International Conference on Multimedia and Expo,
  {ICME}, Hong Kong, China, Jul 10-14,}, 2017, pp. 583--588.

\bibitem{Luo2018}
Danqing Luo, Yuexian Zou, and Dongyan Huang,
\newblock ``Investigation on joint representation learning for robust feature
  extraction in speech emotion recognition,''
\newblock in {\em Interspeech, Hyderabad, India, 2-6 Sep}, 2018, pp. 152--156.

\bibitem{Desplanques2018}
Brecht Desplanques and Kris Demuynck,
\newblock ``Cross-lingual speech emotion recognition through factor analysis,''
\newblock in {\em Interspeech, Hyderabad, India, 2-6 Sep}, 2018, pp.
  3648--3652.

\bibitem{Hermansky1998}
Hynek Hermansky and Sangita Sharma,
\newblock ``{TRAPS} - classifiers of temporal patterns,''
\newblock in {\em International Conference on Spoken Language Processing,
  Australian International Speech Science and Technology Conference, Sydney
  Convention Centre, Sydney, Australia, Nov 30 - Dec 4,}, 1998.

\bibitem{Degottex2014}
Gilles Degottex, John Kane, Thomas Drugman, Tuomo Raitio, and Stefan Scherer,
\newblock ``{COVAREP} - {A} collaborative voice analysis repository for speech
  technologies,''
\newblock in {\em {ICASSP}, Florence, Italy, May 4-9}, 2014, pp. 960--964.

\bibitem{htkbook}
Steve~J. Young, D.~Kershaw, J.~Odell, D.~Ollason, V.~Valtchev, and P.~Woodland,
\newblock {\em {The HTK Book Version 3.4}},
\newblock Cambridge University Press, 2006.

\end{thebibliography}

\end{document}